\documentclass[11pt,reqno]{article}
\usepackage{geometry} 
\usepackage{graphicx}
\usepackage{amsmath}
\usepackage{amsfonts}
\usepackage{amssymb}
\usepackage{xcolor}
\graphicspath{{./figs/}}

\newcommand{\eref}[1]{{Eq.}~\ref{#1}}
\newcommand{\ie}{{\it i.e.}\!\, }

\newcommand{\apriori}{{\it a priori} }

\newcommand{\gb}{\mathbf{g}}
\newcommand{\fb}{\mathbf{f}}
\newcommand{\xb}{\mathbf{x}}
\newcommand{\Ab}{\mathbf{A}}
\newcommand{\Bb}{\mathbf{B}}
\newcommand{\Fb}{\mathbf{F}}
\newcommand{\Ib}{\mathbf{I}}
\newcommand{\Qb}{\mathbf{Q}}
\newcommand{\Rb}{\mathbf{R}}
\newcommand{\Sb}{\mathbf{S}}
\newcommand{\Vb}{\mathbf{V}}
\newcommand{\Xb}{\mathbf{X}}
\newcommand{\Cb}{\mathbf{C}}
\newcommand{\Gb}{\mathbf{G}}
\newcommand{\ws}{\mathsf{w}}
\newcommand{\ys}{\mathsf{y}}
\newcommand{\fs}{\mathsf{f}}
\newcommand{\Is}{\mathsf{I}}
\newcommand{\Ks}{\mathsf{K}}
\newcommand{\Xs}{\mathsf{X}}
\newcommand{\Ibb}{\mathbb{I}}
\newcommand{\Ebb}{\mathbb{E}}
\newcommand{\Vbb}{\mathbb{V}}
\newcommand{\sigmab}{{\boldsymbol{\sigma}}}
\newcommand{\Lambdab}{\boldsymbol{\Lambda}}
\newcommand{\tr}{\operatorname{tr}}
\newcommand{\Cov}{\operatorname{Cov}}
\newcommand{\prob}{p}

\title{\bf Tensor Basis Gaussian Process Models of Hyperelastic Materials}
\author{Ari L. Frankel, Reese E. Jones, Laura P. Swiler\\
\it Sandia National Laboratories
}

\begin{document}

\maketitle
\date{}

\begin{abstract}
In this work, we develop Gaussian process regression (GPR) models of hyperelastic material behavior.
First, we consider the direct approach of modeling the components of the Cauchy stress tensor as a function of the components of the Finger stretch tensor in a Gaussian process.
We then consider an improvement on this approach that embeds rotational invariance of the stress-stretch constitutive relation in the GPR representation.
This approach requires fewer training examples and achieves higher accuracy while maintaining invariance to rotations exactly.
Finally, we consider an approach that recovers the strain-energy density function and derives the stress tensor from this potential.
Although the error of this model for predicting the stress tensor is higher, the strain-energy density is recovered with high accuracy from limited training data.
The approaches presented here are examples of physics-informed machine learning.  
They go beyond purely data-driven approaches by embedding the physical system constraints directly into the Gaussian process representation of materials models.
\end{abstract}

%%%%%%%%%%%%%%%%%%%%%%%%%%%%%%%%%%%%%%%%%%%%%%%%%%%%%%%%%%%%%%%%%%%%%%%%%%%
\section{Introduction}
%%%%%%%%%%%%%%%%%%%%%%%%%%%%%%%%%%%%%%%%%%%%%%%%%%%%%%%%%%%%%%%%%%%%%%%%%%%

Machine learning models have seen an explosion in development and application in recent years due to their flexibility and capacity for capturing the trends in complex systems \cite{hastie2016elements}.
Provided with sufficient data, the parameters of the model may be calibrated in such a way that the model gives high fidelity representations of the underlying data generating process \cite{raissi2017,jones2018machine,frankel2019predicting,frankel2019prediction}.
Moreover, computational capabilities have grown such that constructing deep learning models over datasets of tens of thousands to  millions of data points is now feasible \cite{dean2012large}.
There remain, however, many applications in which the amount of data present is insufficient on its own to properly train the machine learning model.
This may be due to a prohibitively large model that requires a correspondingly large amount of data to train and where training data is expensive to acquire.
Furthermore, even with a wealth of data, it is possible that the machine learning model may yield behavior that is inconsistent with the expected trend of the model when the model is queried in an extrapolatory regime.

In such cases it is appealing to turn to a framework that allows the incorporation of physical principles and other \apriori information to supplement the limited data and regularize the behavior of the model.
This information can be as simple as a known set of constraints that the regressor must satisfy, such as positivity or monotonicity with respect to a particular variable, or can be as complex as knowledge of the underlying data-generating process in the form of a partial differential equation.
Consequently, the past few years have seen great interest in ``physics-constrained'' machine learning algorithms within the scientific computing community \cite{raissi2018, pan2018data, lusch2018deep, brunton2016discovering, LeeCarlberg, ling2016machine,jones2018machine}. 
The overview paper by Karpatne et al. \cite{karpatne2017theory} provides a taxonomy for theory-guided data science, with the goal of incorporating scientific
consistency in the learning of generalizable models.     
Much research in physics-informed machine learning has focused on incorporating constraints in neural networks \cite{ling2016machine,jones2018machine}, often through the use of objective/loss functions which penalize constraint violation \cite{magiera2019constraint}.

In contrast, the focus in this paper is to incorporate rotational symmetries directly and exactly into  Gaussian process representations of physical response functions.
This approach has the advantages of avoiding the burden of a large training set that comes with neural network model, and the inexact satisfaction of constraints that come with penalization of constraints in the loss function.
 There has been significant interest in the incorporation of constraints into Gaussian process regression models recently \cite{riihimaki, daveiga, jensen, solak, raissi2017, yang, lopez2018, bachoc2019}.
Many of these approaches leverage the analytic formulation of the Gaussian process (GP) to incorporate constraints through the likelihood function or the covariance function.
 
In this paper, the task of learning the 6 components of a symmetric stress tensor from the 6 components of a symmetric stretch tensor is formulated through a series of transformations so that it becomes a regression task of learning three coefficients that are a function of three invariants of the problem.
The main contribution of this paper is the extension of Gaussian process regression to enforce rotational invariance through a tensor basis expansion.
 
The paper is organized as follows: Section 2 presents an overview of constitutive models for hyperelastic materials.
Sections 3 and 4 present Gaussian process regression and the extension to a tensor basis Gaussian process, respectively.
Section 5 presents a further extension of the tensor-basis GP to handle the strain energy potential.
Section 6 provides results for a particular hyperelastic Mooney-Rivlin material, and Section 7 provides concluding discussion.

%%%%%%%%%%%%%%%%%%%%%%%%%%%%%%%%%%%%%%%%%%%%%%%%%%%%%%%%%%%%%%%%%%%%%%%%%%%
\section{Hyperelastic Materials}
%%%%%%%%%%%%%%%%%%%%%%%%%%%%%%%%%%%%%%%%%%%%%%%%%%%%%%%%%%%%%%%%%%%%%%%%%%%

A hyperelastic material is a material that remains elastic (non-dissipative) in the finite/large strain regime.
In this context the fundamental deformation measure is the $3 \times 3$ deformation gradient tensor $\Fb$: 
\begin{equation}
\Fb = \frac{\partial \xb}{\partial \Xb} \ ,
\end{equation}
which is the derivative of the current position $\xb$ with respect to position $\Xb$ of the same material point in a chosen reference configuration.
In an Eulerian frame, the Finger tensor $\Bb$
\begin{equation}
\Bb = \Fb \Fb^T \ .
\end{equation}
is the typical finite stretch measure, which is directly related to the Almansi strain, which measures the total deformation that a material has undergone relative to its initial configuration.
(The choice of the Finger tensor is not limiting in terms of the generality of this formulation, given the equivalence of strain measures provided by the Seth-Hill \cite{hill1968constitutive}/Doyle-Ericksen\cite{doyle1956nonlinear} formulae.)
The deformation of a hyperelastic material requires an applied stress state, associated with a certain amount of energy, to arrive at that deformed state. For a hyperelastic material, the stress is solely a function of the current stretch (or strain) of the material.
Hence, the major goal of material modeling of hyperelastic materials is to construct constitutive relations between the kinematic variable $\Bb$ and the corresponding dynamic variable, the $3 \times 3$ Cauchy stress tensor
\begin{equation} \label{eq:potential}
\sigmab = \fb(\Bb) 
= \frac{2}{I_3^{1/2}}\Bb\frac{\partial \Phi}{\partial \Bb}
\end{equation}
which for a hyperelastic material is given by the derivative of a potential, namely the strain energy density $\Phi$. 
For further details please consult Refs. \cite{malvern1969introduction,ogden1997non,gurtin1982introduction}.

Typical approaches to model these relations seek semi-empirical formulations for the strain energy density with some parameters to be fit, which are then fit to experimental data. An example of this type of formulation will be discussed in a later section. In this work we consider non-parametric modeling of hyperelastic material responses.

%%%%%%%%%%%%%%%%%%%%%%%%%%%%%%%%%%%%%%%%%%%%%%%%%%%%%%%%%%%%%%%%%%%%%%%%%%%
\section{Gaussian Process Regression}
%%%%%%%%%%%%%%%%%%%%%%%%%%%%%%%%%%%%%%%%%%%%%%%%%%%%%%%%%%%%%%%%%%%%%%%%%%%

Gaussian process regression (GPR) provides a non-parametric model for a response function given an input set of training data through a Bayesian update involving an assumed prior distribution and a likelihood tying the posterior distribution to observed data.
We denote a Gaussian process prior for a function $f$ by
\begin{equation}
f \sim \mathcal{G\!P}(0,K) \ ,
\end{equation}
where we assume the GP has a nominal mean of $0$, without loss of generality, and is described by a covariance function $K$.
We adopt the commonly employed squared-exponential covariance function:
\begin{equation} \label{eq:kernel}
K(x,x') = \theta_1 \exp(-\theta_2 |x-x'|^2)
\end{equation}
which has a scale parameter $\theta_1$ and a length parameter $\theta_2$.

A GP is defined such that any finite collection of realizations from the process are governed by a multivariate normal distribution.
That is, for any set of observed realizations $\Xs$ and prediction points $\Xs^*$ with corresponding function values $f(\Xs)$ and $f(\Xs^*)$, the probability distribution $\prob(f(\Xs^*),f(\Xs)|\Xs^*,\Xs)$ is given by
\begin{equation}
\begin{bmatrix}
f(\Xs)\\
f(\Xs^*)
\end{bmatrix}
=
 \mathcal{N}\left(\begin{bmatrix}
\mathbf{0}\\
\mathbf{0}
\end{bmatrix}
,
\begin{bmatrix}
K(\Xs,\Xs) & K(\Xs,\Xs^*)\\
K(\Xs^*,\Xs) & K(\Xs^*,\Xs^*)
\end{bmatrix}
\right)
\end{equation}
where it is understood that the vectors and matrices presented are given in block form for multiple instances in $\Xs$ and $\Xs^*$. Gaussian process regression (GPR) uses a GP as a prior over the function space for the data-generating process, and predictions proceed through the use of Bayes' rule.
Upon observation of some initial set of noisy data points $\ys = y(\Xs)=f(\Xs)+\varepsilon$ with Gaussian noise of variance $\varepsilon^2$ in the function values, the probability distribution of the values of the GP at $\Xs^*$ may be determined by forming the (posterior) conditional distribution of $\prob(\fs^*|\ys,\Xs)$:
\begin{equation}
\prob(\fs^*|\ys,\Xs) =\frac{\prob(\ys|\fs)\prob(\fs|\Xs)}{\prob(\ys|\Xs)} 
= \frac{\prob(\ys|\fs)\prob(\fs|\Xs)}{\int \prob(\ys|\fs)\prob(\fs|\Xs)\mathrm{d}\fs}
\end{equation}
where $\fs = f(\Xs)$ and the Gaussian likelihood is given by
\begin{equation}
\prob(\ys|\fs) = \prod_{i=1}^N \frac{1}{\sqrt{2\pi\varepsilon^2}}\exp\left(-\frac{(y_i-f_i)^2}{2\varepsilon^2}\right) \ .
\end{equation}
This posterior distribution has the following analytical solution for the underlying mean process:
\begin{equation}
\fs^* = f(\Xs^*) \sim \mathcal{N}(\Ks^* (\Ks+\varepsilon^2 \Is)^{-1}\ys, \,
\Ks^{**}-\Ks^*(\Ks+\varepsilon^2 \Is)^{-1}\Ks^{*T})
\end{equation}
where $\Is$ denotes the identity matrix, $\Ks=K(\Xs,\Xs)$, $\Ks^* = K(\Xs^*,\Xs)$, and $\Ks^{**}= K(\Xs^*,\Xs^*)$. 
Then the predictive mean of the distribution at any new points $\Xs^*$ is given by
\begin{equation} \label{eq:pred_mean}
\Ebb[\ys^*] = \Ks^*(\Ks+\varepsilon^2 \Is)^{-1}\ys
\end{equation}
and the predictive variance, assuming the same noise level, is given by
\begin{equation} \label{eq:pred_var}
\Vbb[\ys^*] = \Ks^{**}-\Ks^*(\Ks+\varepsilon^2 \Is)^{-1} \Ks^{*T} + \varepsilon^2 \Is
\end{equation}
where $\ys^* = y(\Xs^*)$. This result shows the combination of uncertainty in the prediction due to epistemic uncertainty in the mean process (the first two terms on the right side) plus the aleatoric uncertainty of inherent variability in the measurements (the last term on the right side). In this work, although we will work with noiseless data, we assume a value of $\varepsilon^2 = 10^{-10}$ in order to regularize the inversion of the covariance matrix.

The task that dominates the computational expense in constructing this model is the inversion of $(\Ks+\varepsilon^2 \Is)$, or, equivalently, the solution of the linear system based on $(\Ks+\varepsilon^2 \Is)$ for either the mean or variance evaluations.
Since $\Ks$ is dense, the scaling is typically $\mathcal{O}(N^3)$ for $N$ training points.
Nominally the matrix is symmetric positive semi-definite, which enables efficient solution by Cholesky decompositions, although ill-conditioning is frequently an issue.
Ill-conditioning requires adding a ridge or large noise ($\varepsilon \gg 1$) term to the covariance matrix to regularize the solution, using pseudoinverses via the singular value decomposition (SVD) of the covariance matrix, or other greedy subset selection to reduce the matrix size.

It appears at first glance that the variance in \eref{eq:pred_var} is nominally independent of the actual point values $\ys$, and only depends on the locations of the selected data points $\Xs$.
This is true for a fixed covariance function; however, we are typically interested in changing the GP hyperparameters to maximize the accuracy of the GP while balancing the model complexity.
Traditionally, this is managed by tuning the hyperparameters to optimize $\prob(\ys|\Xs)$, which is the marginal-likelihood of the GP, and is frequently called the ``model evidence.'' 
Equivalently, we may optimize the logarithm of the model evidence $L=\log \prob(\ys|\Xs)$ for numerical stability reasons:
\begin{equation}
L = -\frac{1}{2} \fs^T (\Ks+\varepsilon^2 \Is)^{-1} \fs - \frac{1}{2}\log |\Ks+\varepsilon^2 \Is| - \frac{N}{2}\log 2\pi
\end{equation}
That is, we choose to tune the covariance hyperparameters $\theta_1$ and $\theta_2$ in \eref{eq:kernel} in order to maximize $L$. 
Further discussion of this approach can be found in Ref. \cite{rasmussen}.

%%%%%%%%%%%%%%%%%%%%%%%%%%%%%%%%%%%%%%%%%%%%%%%%%%%%%%%%%%%%%%%%%%%%%%%%%%%
\section{Tensor Basis Gaussian Process}
%%%%%%%%%%%%%%%%%%%%%%%%%%%%%%%%%%%%%%%%%%%%%%%%%%%%%%%%%%%%%%%%%%%%%%%%%%%

In this section we show how the standard Gaussian process regression described in the previous section may be adapted to enforce rotational invariance through a tensor basis expansion.
We call this formulation a tensor basis Gaussian process (TBGP).

%============================================================================
\subsection{Tensor Basis Expansion}
%============================================================================

We consider the generic hyperelastic constitutive model of the form
\begin{equation}
\sigmab = \fb(\Bb)
\end{equation}
which, for any given continuous and differentiable tensor valued function $\fb$, may be expanded in an infinite series in terms of $\Bb$ with fixed coefficients $\bar{c}_n$
\begin{equation}
\sigmab = \sum_{n=0}^\infty \bar{c}_n \Bb^n
\end{equation}
It is clear that $\sigmab$ and $\Bb$ are collinear, \ie  have the same eigenbasis.
Since the tensors of interest are symmetric and of size 3$\times$3, the Cayley-Hamilton theorem states that the tensor $\Bb$ satisfies its corresponding characteristic polynomial
\begin{equation}
\Bb^3 - I_1 \Bb^2 + I_2 \Bb - I_3 \Ib = \mathbf{0}
\end{equation}
where we have defined the tensor invariants
\begin{alignat}{3}
I_1 &= \tr(\Bb) 
    &&= \lambda_{B_1} + \lambda_{B_2} + \lambda_{B_3}, \nonumber \\
I_2 &= \frac{1}{2}(\tr(\Bb)^2-\tr(\Bb^2)) 
    &&= \lambda_{B_1} \lambda_{B_2} + \lambda_{B_2} \lambda_{B_3} + \lambda_{B_3} \lambda_{B_1}, \label{eq:invariants} \\
I_3 &= \det(\Bb)
    &&= \lambda_{B_1} \lambda_{B_2} \lambda_{B_3}, \nonumber
\end{alignat}
so-called because they are invariant under similarity transformations (\ie rotations) of $\Bb$.
Here, $\lambda_{B_1}$, $\lambda_{B_2}$, $\lambda_{B_3}$ are the eigenvalues of $\Bb$, which are also a complete set of invariants.
The theorem can be used as a recursion relation to write all powers of $\Bb$ higher than $2$ in terms of $\Ib$, $\Bb$, and $\Bb^2$ with coefficients that depend on the invariants.
Rather than seeking to identify the infinite number of fixed coefficients $\bar{c}_i$ for a given constitutive relation, our task reduces to finding the three coefficients in the series expansion 
\begin{equation} \label{eq:expansion}
\sigmab = c_1 \Ib + c_2 \Bb + c_3 \Bb^2 , 
\end{equation}
where $c_i$ is a function of the invariants.

This reduced expansion maintains rotational objectivity for the original functional dependence for the appropriately defined coefficients.
To see this, let $\Rb$ be an orthogonal/rotation tensor with inverse given by $\Rb^{-1} = \Rb^T$.
The rotation of $\sigmab$ in the original coordinate frame to the frame defined by $\Rb$ is given by
\begin{equation}
\sigmab' \equiv \Rb \sigmab \Rb^T = \Rb \fb(\Bb) \Rb^T
\end{equation}
Invoking the tensor basis expansion gives
\begin{align}
\sigmab' &= c_1 \Rb \Ib \Rb^T + c_2 \Rb \Bb \Rb^T + c_3 \Rb \Bb^2 \Rb^T  \\
         &= c_1 \Rb \Rb^T + c_2 \Rb \Bb \Rb^T + c_3 \Rb \Bb \Rb^T \Rb \Bb \Rb^T = \fb(\Rb\Bb\Rb^T) \equiv \fb(\Bb') \nonumber
\end{align}
which holds since the eigenvalues, and hence invariants and the coefficient functions, do not change upon application of $\Rb$.
In general, an Eulerian tensor function of an Eulerian tensor argument must be objective in the sense it responds to a rotation of its argument with a corresponding rotation of the function value:
\begin{equation}
\Rb \fb(\Bb) \Rb^T  = \fb(\Rb \Bb \Rb^T) \ .
\end{equation}

%============================================================================
\subsection{Application to Gaussian Process Modeling} \label{sec:app}
%============================================================================

The task of regression now falls to learning the coefficients $c_1$, $c_2$, $c_3$ as a function of the invariants.
This task is compressed from the original problem of having to learn 6 stress components from 6 strain components with the added benefit of enforcing the rotational invariance that provides this reduction.

Suppose we have been given a dataset of pairs of tensors ($\Bb$, $\sigmab$).
Under the assumption that the tensors may be diagonalized, the collinearity of the tensor basis expansion \eref{eq:expansion} implies that they are diagonalized by the same eigenvector matrix $\Qb$ but with different eigenvalue matrices, $\Lambdab_\sigmab$ and $\Lambdab_\Bb$, respectively:
\begin{equation}
\sigmab = \Qb\Lambdab_\sigmab \Qb^T
\end{equation}
\begin{equation}
\Bb = \Qb\Lambdab_\Bb \Qb^T.
\end{equation}
Then for the given input-output pair, the values of the coefficients for the given set of eigenvalues are given by the solution to a 3$\times$3 linear system of equations as
\begin{equation} \label{eq:system}
\begin{bmatrix}
\lambda_{\sigma_1} \\ \lambda_{\sigma_2} \\ \lambda_{\sigma_3}
\end{bmatrix}
=
\begin{bmatrix}
1 & \lambda_{B_1} & \lambda_{B_1}^2\\
1 & \lambda_{B_2} & \lambda_{B_2}^2\\
1 & \lambda_{B_3} & \lambda_{B_3}^2\\
\end{bmatrix}
\begin{bmatrix}
c_1 \\ c_2 \\ c_3
\end{bmatrix}.
\end{equation}
This linear system may be inverted easily to yield the coefficients $c_i(\Bb,\sigmab)$ for each training point in the dataset, and the invariants $I_i$ of $\Bb$ may also be computed straightforwardly from the eigenvalues given in \eref{eq:invariants}.
The three invariants $I_i$ for each observation are accumulated into a matrix.
They take the place of the feature matrix $\Xs$ presented in the previous section and a corresponding matrix of $c_i$ replaces $\fs$.
We can thus readily extend the GPR approach to infer the coefficients as functions of the invariants of the input tensors and thus construct the representation of the function $\sigmab = c_1 \Ib + c_2 \Bb + c_3 \Bb^2$ given in \eref{eq:expansion}.

%============================================================================
\subsection{Example: Matrix Exponential}
%============================================================================

To illustrate the impact of embedding rotational invariance in the GP formulation, we consider the representation of the matrix exponential
\begin{equation}
\Sb = \exp(\Bb) = \Qb\exp(\Lambdab)\Qb^T
\end{equation}
from a limited number of training samples.
As in Sec.~\ref{sec:app}, $\Bb=\Qb\Lambdab\Qb^T$ is a symmetric diagonalizable matrix with eigenvector matrix $\Qb$ and diagonal eigenvalue matrix $\Lambdab$. 
Clearly the matrix exponential maintains rotational invariance under application of a rotation $\Rb$  and the series representation of $\exp(\Bb)$ demonstrates the collinearity of  $\exp(\Bb)$ and $\Bb$.
Given \eref{eq:system}, the expansion coefficients may be determined from
\begin{equation}
\begin{bmatrix}
\exp{\lambda_{1}} \\ 
\exp{\lambda_{2}} \\ 
\exp{\lambda_{3}}
\end{bmatrix}
=
\begin{bmatrix}
1 & \lambda_{1} & \lambda_{1}^2\\
1 & \lambda_{2} & \lambda_{2}^2\\
1 & \lambda_{3} & \lambda_{3}^2\\
\end{bmatrix}
\begin{bmatrix}
c_1 \\ c_2 \\ c_3
\end{bmatrix}
\end{equation}

To create the dataset, we draw random uniformly distributed 3$\times$3 matrices with entries between $[0,1]$ and compute their symmetric part.
A single GP is formed over the 6 independent tensor components (the upper triangular part) of $\Bb$ to predict the 6 independent output tensor components.
This formulation is compared against the TBGP formulation, where the 3 invariants are used to predict the 3 expansion coefficients with a single GP. 
The {\it scikit-learn} library \cite{sklearn} was used to train the GPs in both cases.
The root-mean-squared error is evaluated in each case at 10,000 testing points for validation.
The input training points are then rotated randomly, and the GP prediction is evaluated at the inputs.

\begin{figure}
\centering
\includegraphics[width=0.5\textwidth]{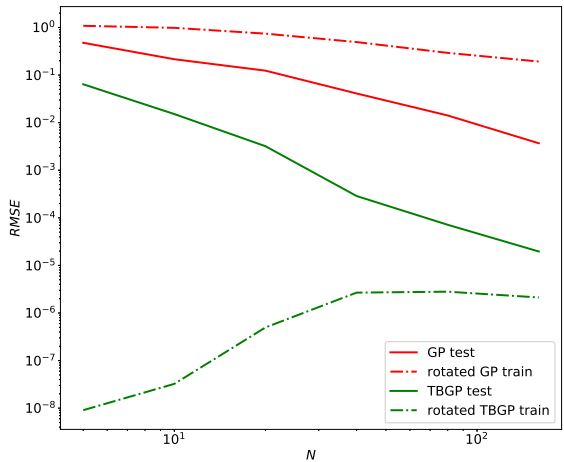}
\caption{The root-mean-squared error in predicting the matrix exponential for the GP and TBGP formulations as a function of training set size.
The results also show the RMSE of the regressors evaluated at random rotations of the training set pairs.
}
\label{expm}
\end{figure}

Figure \ref{expm} shows the results of the testing error as a function of increasing training points for the GP and TBGP formulations.
Since the rotationally invariant formulation does not take the orientation of the eigenvectors into account, it is expected that the prediction error on the modified inputs in this case would be small, whereas the prediction error for the normal GP could be quite large.
The TBGP error is indeed 1-2 orders of magnitude lower than the GP error on the testing sets, and the error on the randomly-rotated training set is also over 5 orders of magnitude lower, demonstrating that the TBGP formulation learns the underlying function much more quickly than a standard GP. It also appears to take an order of magnitude order more data before the GP error catches up to the error that the TBGP had on the smallest dataset. 
The error on the rotated training set appears to increase, which we attribute to a combination of increasing condition number of the covariance matrix and accumulated interpolation error in the GP through the training points from the use of regularization to maintain invertibility.
Even with these effects, the TBGP has very little error throughout the tests and is uniformly the better choice.

%%%%%%%%%%%%%%%%%%%%%%%%%%%%%%%%%%%%%%%%%%%%%%%%%%%%%%%%%%%%%%%%%%%%%%%%%%%
\section{Strain Energy Potential Gaussian Process}
%%%%%%%%%%%%%%%%%%%%%%%%%%%%%%%%%%%%%%%%%%%%%%%%%%%%%%%%%%%%%%%%%%%%%%%%%%%

The tensor basis GP formulation is quite powerful and could be made general to many different types of processes.
For hyperelastic materials, where the stress is the derivative of a potential, it is possible to take this approach one step further with an alternate formulation.
As in \eref{eq:potential}, we define the strain-energy density function $\Phi$ such that the stress tensor may be computed as
\begin{equation}
\sigmab = \frac{2}{I_3^{1/2}}\Bb\frac{\partial \Phi}{\partial \Bb}
\end{equation}
for some appropriate $\Phi(\Bb)$.
The strain-energy density is an invariant scalar quantity and cannot depend on the rotation of the frame; thus for an isotropic hyperelastic material it is preferable to express the strain-energy density as a function of the invariants of $\Bb$.
Thus the expression for the stress tensor may be expanded as
\begin{equation}
\sigmab = \frac{2}{I_3^{1/2}}\Bb\left( \frac{\partial \Phi}{\partial I_1}\frac{\partial I_1}{\partial \Bb} + \frac{\partial \Phi}{\partial I_2}\frac{\partial I_2}{\partial \Bb} + \frac{\partial \Phi}{\partial I_3}\frac{\partial I_3}{\partial \Bb}\right)
\end{equation}
where we have applied the chain rule for the strain-energy density partial derivatives.
The derivatives of the invariants with respect to the original tensor may be computed directly, refer to Ref. \cite{bonet1997nonlinear}, and the expression for $\sigmab$ simplifies to
\begin{equation}
\sigmab = \frac{2}{I_3^{1/2}}\left(I_3 \frac{\partial \Phi}{\partial I_3} \Ib + \left(\frac{\partial \Phi}{\partial I_1}+I_1\frac{\partial \Phi}{\partial I_2}\right)\Bb - \frac{\partial \Phi}{\partial I_2}\Bb^2\right).
\end{equation}
This expression explicitly takes the form of a tensor-basis expansion of the type in \eref{eq:expansion}, where we have
\begin{align}
c_1 &= 2I_3^{1/2}\frac{\partial \Phi}{\partial I_3} \nonumber \\
c_2 &= \frac{2}{I_3^{1/2}}\left(\frac{\partial \Phi}{\partial I_1}+I_1\frac{\partial \Phi}{\partial I_2}\right) \\
c_3 &= -\frac{2}{I_3^{1/2}}\frac{\partial \Phi}{\partial I_2} \nonumber
\end{align}

For a given set of coefficients and invariants, the corresponding partial derivatives can be evaluated using the following relations:
\begin{align} 
\frac{\partial \Phi}{\partial I_1} &= \frac{I_3^{1/2}}{2}(c_2 + c_3 I_1) \nonumber \\
\frac{\partial \Phi}{\partial I_2} &= -\frac{c_3 I_3^{1/2}}{2} \label{eq:dPhidI} \\
\frac{\partial \Phi}{\partial I_3} &= \frac{c_1}{2I_3^{1/2}} \nonumber
\end{align}
Thus given a set of observed pairs of $(\sigmab,\Bb)$ and Eqs. \ref{eq:system} and \ref{eq:dPhidI}, we can infer the corresponding values of the gradient of the strain-energy density function.
Furthermore, we ``ground'' the function  (\ie remove the indeterminacy of calibration of $\Phi$ from derivative data) by choosing a zero-point energy for a set of invariants where the material has not been deformed.
Hence, we augment the gradient information with the datum
\begin{equation}
\Phi(I_1=3,I_2=3,I_3=1) = 0 \ .
\end{equation}

The GP regression technique can be extended to take advantage of derivative information since the derivative of a GP is also a GP.
Specifically, the covariance between the derivatives of the stress-energy density between two different points in invariant space, $\Ibb =  (I_1,I_2,I_3)$ and $\Ibb'= (I'_1,I'_2,I'_3)$, can be evaluated as
\begin{align}
\Cov\left(\frac{\partial \Phi(\Ibb)}{\partial I_i},\Phi(\Ibb')\right) &= \frac{\partial K(\Ibb,\Ibb')}{\partial I_i} \\
\Cov\left(\frac{\partial \Phi(\Ibb)}{\partial I_i},\frac{\partial \Phi(\Ibb')}{\partial I'_j} \right) &= \frac{\partial^2 K(\Ibb,\Ibb')}{\partial I_i \partial I'_j}
\end{align}
Using these relations, a GP may be formed simultaneously over $\Phi$ at the grounding point and its derivatives over the dataset by using the block covariance matrix 
\begin{equation} \label{eq:Kp}
\Ks_\Phi(\Ibb,\Ibb') = 
\begin{bmatrix}
\frac{\partial^2 K}{\partial I_1 \partial I'_1}  & \frac{\partial^2 K}{\partial I_1 \partial I'_2} & \frac{\partial^2 K}{\partial I_1\partial I'_3}\\
\frac{\partial^2 K}{\partial I_2 \partial I'_1}  & \frac{\partial^2 K}{\partial I_2 \partial I'_2} & \frac{\partial^2 K}{\partial I_2\partial I'_3}\\
\frac{\partial^2 K}{\partial I_3 \partial I'_1}  & \frac{\partial^2 K}{\partial I_3 \partial I'_2} & \frac{\partial^2 K}{\partial I_3\partial I'_3}\\
\end{bmatrix}
\end{equation}
where each entry is a matrix corresponding to the covariance between the individual derivatives of $\Phi$ evaluated between each of the training data points.
The grounding point $\Ibb_g = (3,3,1)$ is included by augmenting this matrix with an additional row and column:
\begin{equation} \label{eq:Kg}
\Ks_g(\Ibb,\Ibb') = 
\begin{bmatrix}
K(\Ibb_g,\Ibb_g) & \frac{\partial K(\Ibb_g,\Ibb')}{\partial I'_1} & \frac{\partial K(\Ibb_g,\Ibb')}{\partial I'_2} & \frac{\partial K(\Ibb_g,\Ibb')}{\partial I'_3}\\
\frac{\partial K(\Ibb,\Ibb_g)}{\partial I_1} & & &\\
\frac{\partial K(\Ibb,\Ibb_g)}{\partial I_2} & &\Ks_\Phi(\Ibb,\Ibb') &\\
\frac{\partial K(\Ibb,\Ibb_g)}{\partial I_3} & & &
\end{bmatrix}
\end{equation}
In a slight abuse of notation, in the matrix $\Ks_\Phi(\Ibb,\Ibb')$ in \eref{eq:Kp} and the matrix $\Ks_g(\Ibb,\Ibb')$ in \eref{eq:Kg}, the arguments $\Ibb$ and $\Ibb'$ should be interpreted as matrices of the invariants at the training points (as opposed to individual points).

The GP mean of the potential at a new point $\Ibb^* = ( I^*_{1}, I^*_{2}, I^*_{3} )$ may then be predicted with
\begin{equation}
\Ebb[\Phi(\Ibb^*)] = \Ks_g(\Ibb^*,\Ibb) \, \ws
\end{equation}
where we have defined the weight vector
\begin{equation}
\ws = \Ks_g^{-1} (\Ibb,\Ibb')
\begin{bmatrix}
0 &
\frac{\partial \Phi}{\partial {I_1}} &
\frac{\partial \Phi}{\partial {I_2}} &
\frac{\partial \Phi}{\partial {I_3}}
\end{bmatrix}^T
\end{equation}
where the right hand side partial derivatives are at the observed data points, and $\Ks_g(\Ibb^*,\Ibb)$ is the covariance between the test points and the training points augmented with the ground point.
This expression is analogous to \eref{eq:pred_mean}, although here we have omitted the noise term.
The gradient of the potential $\Phi$, and hence the stress $\sigmab$, may be evaluated using the same weight vector.

\section{Results: Mooney-Rivlin Material}
%%%%%%%%%%%%%%%%%%%%%%%%%%%%%%%%%%%%%%%%%%%%%%%%%%%%%%%%%%%%%%%%%%%%%%%%%%%

In this section we consider the application of GPR to predicting data drawn from the stress response of the deformation of a hyperelastic material.
The underlying truth model will be assumed to be a compressible Mooney-Rivlin material with strain-energy density function
\begin{equation}
\Phi = c_1 (I_3^{-1/2} I_1 -3) + c_2 (I_3^{-2/3}I_2-3) + c_3 (I_3^{1/2}-1)^2
\end{equation}
where we take $c_1 = 0.162$ MPa, $c_2 = 0.0059$ MPa, and $c_3 = 10$ MPa \cite{marckmann2006comparison} and we make $c_3 \gg c_1, c_2$ large to effect nearly incompressible response.

We generate realizations of arbitrary mixed compression/tension/shear states using the following procedure.
Let $\Vb$ be a diagonal matrix of randomly sampled positive values between $0<l\leq1<u$, and let $\Rb$ be a random rotation matrix sampled uniformly on SO(3).
We then employ the polar decomposition of the deformation gradient tensor 
\begin{equation}
\Fb = \Rb \Vb
\end{equation}
with corresponding Finger tensor
\begin{equation}
\Bb = \Rb \Vb^2 \Rb^T
\end{equation}
thus guaranteeing that the determinant of $\Fb$ is positive and that $\Bb$ corresponds to a valid diagonalizable tensor with some superposition of tension/compression and shear in arbitrary directions. The corresponding eigenvalues of $\Bb$ are randomly distributed in the interval $[l^2,u^2]$.

\begin{figure}
\centering
\includegraphics[width=0.5\textwidth]{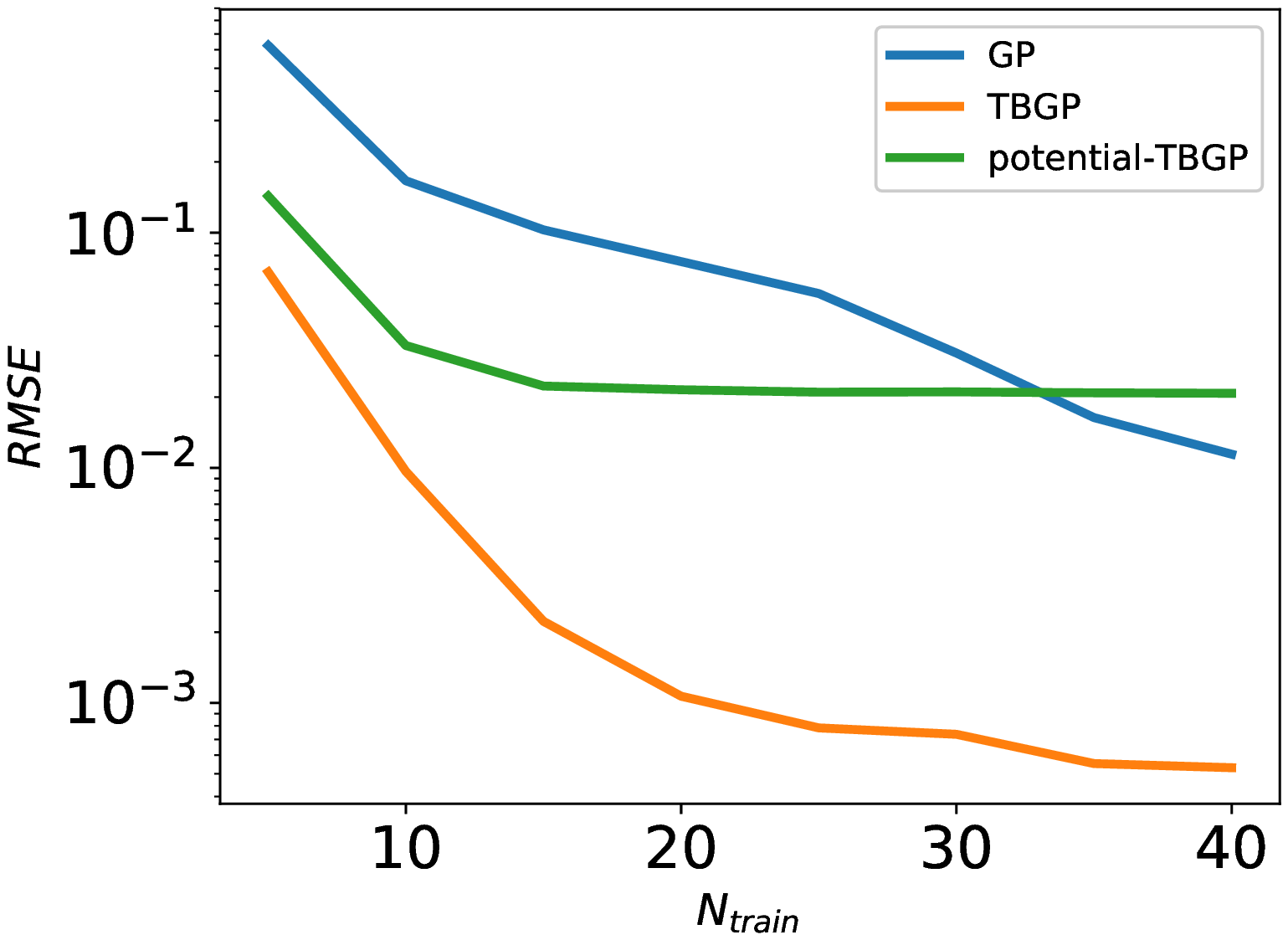}
\includegraphics[width=0.5\textwidth]{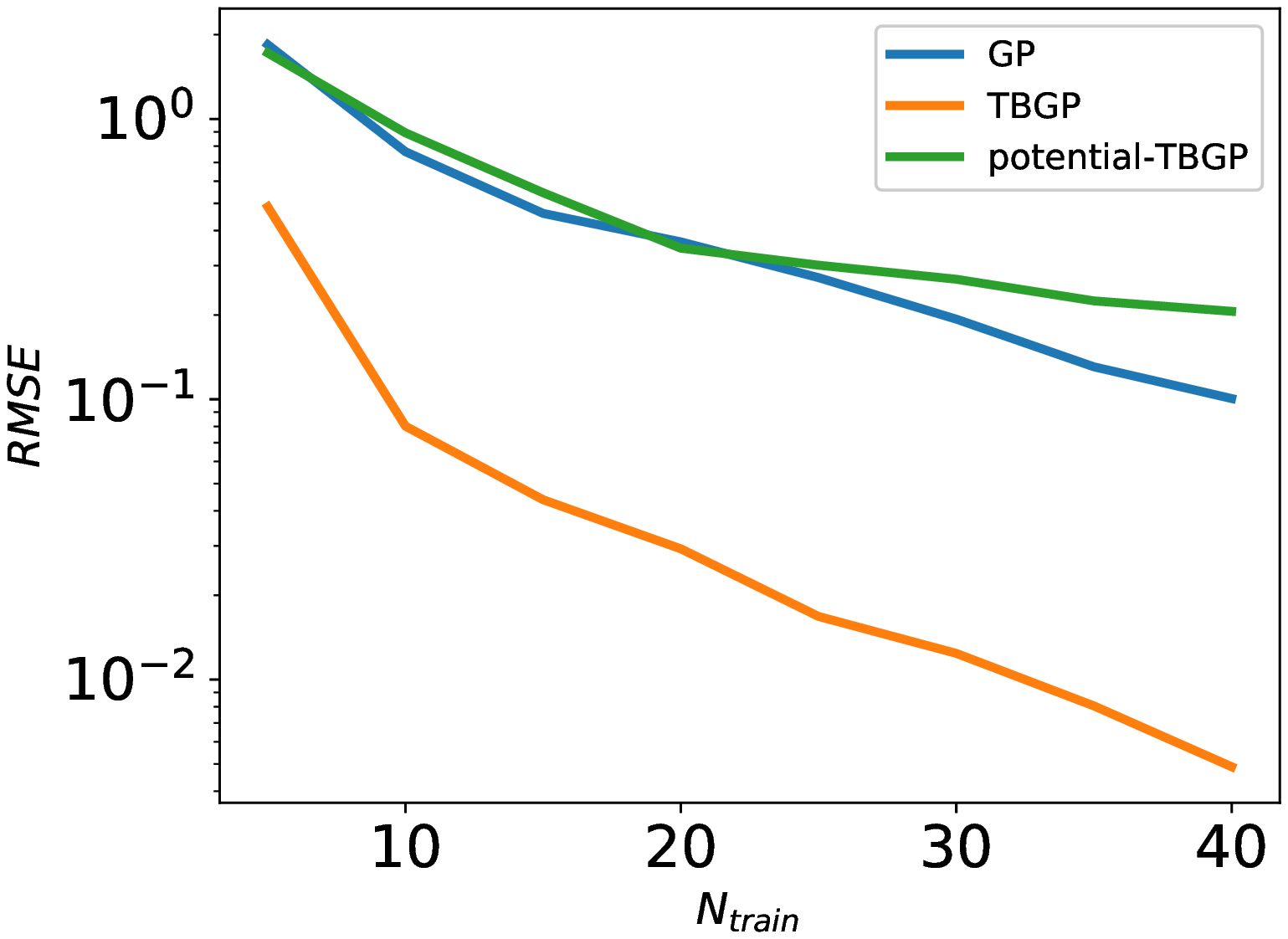}
\caption{The root-mean-squared error of the stress tensor for the GP, TBGP, and potential-TBGP formulations as a function of training set size for $[l^2,u^2] = [1.0,1.5]$ (upper panel) and $[0.9,2.0]$ (lower panel).
}
\label{fig:stress_rmse}
\end{figure}

\begin{figure}
\centering
\includegraphics[width=0.95\textwidth]{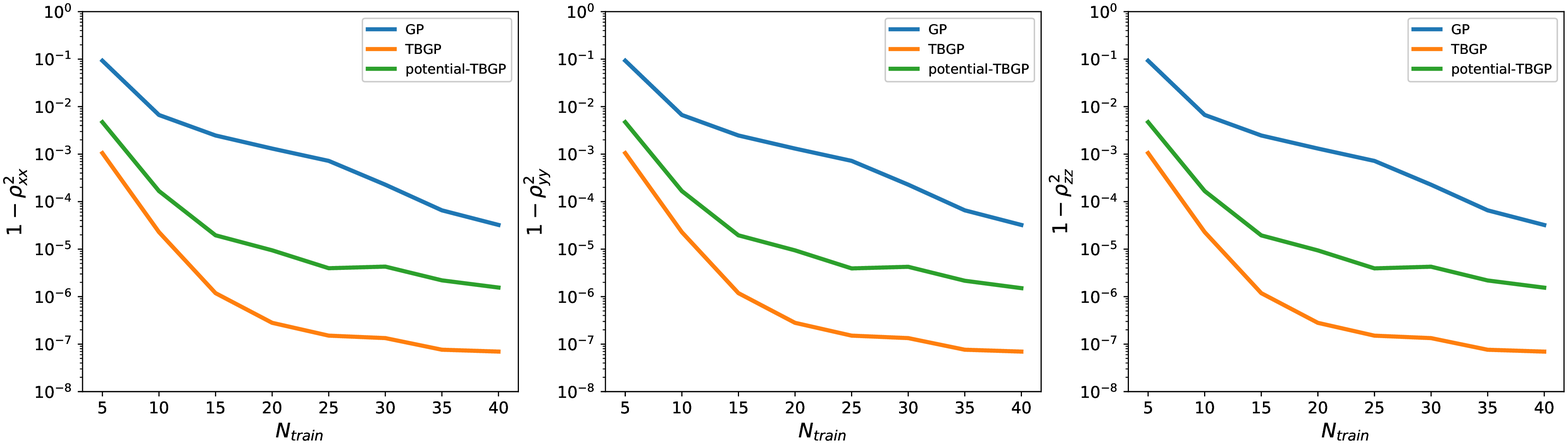}
\includegraphics[width=0.95\textwidth]{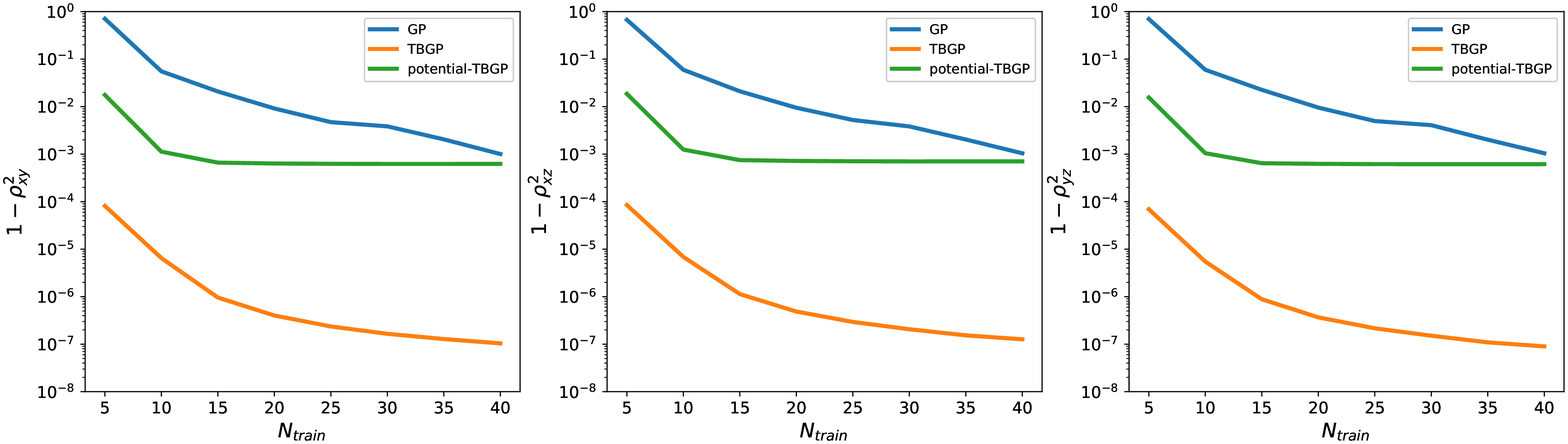}
\caption{The component-wise fraction of variance unexplained ($1-\rho^2$) for the stress tensor predictions as a function of training set size for $[l^2,u^2] = [1.0,1.5]$. Top row are the tension components, and the bottom are the shear components.
}
\label{fig:stress_corr2}
\end{figure}

\begin{figure}
\centering
\includegraphics[width=0.95\textwidth]{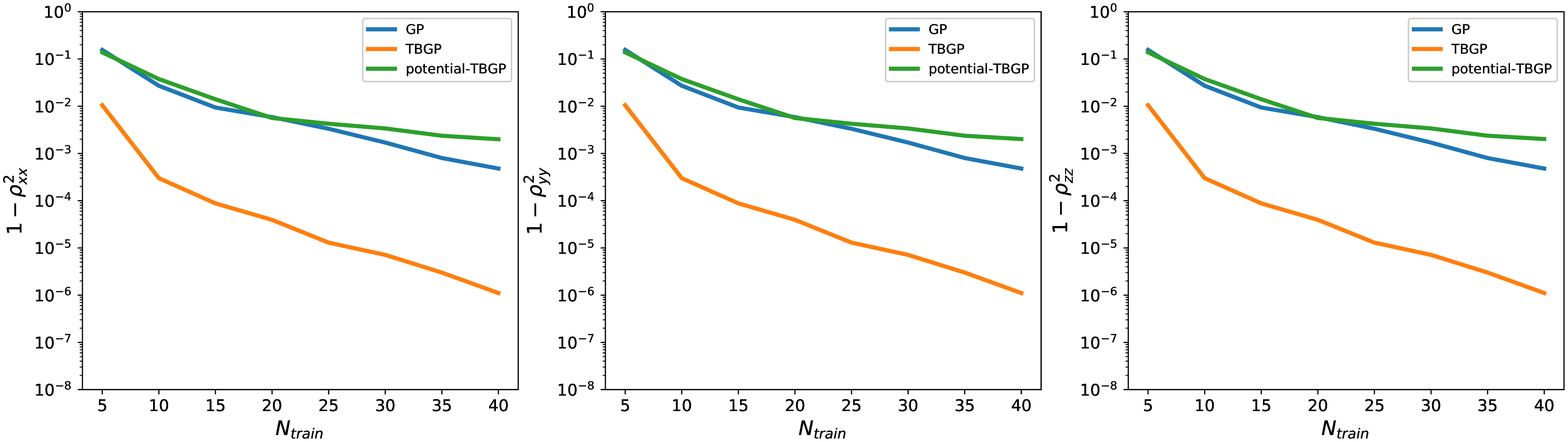}
\includegraphics[width=0.95\textwidth]{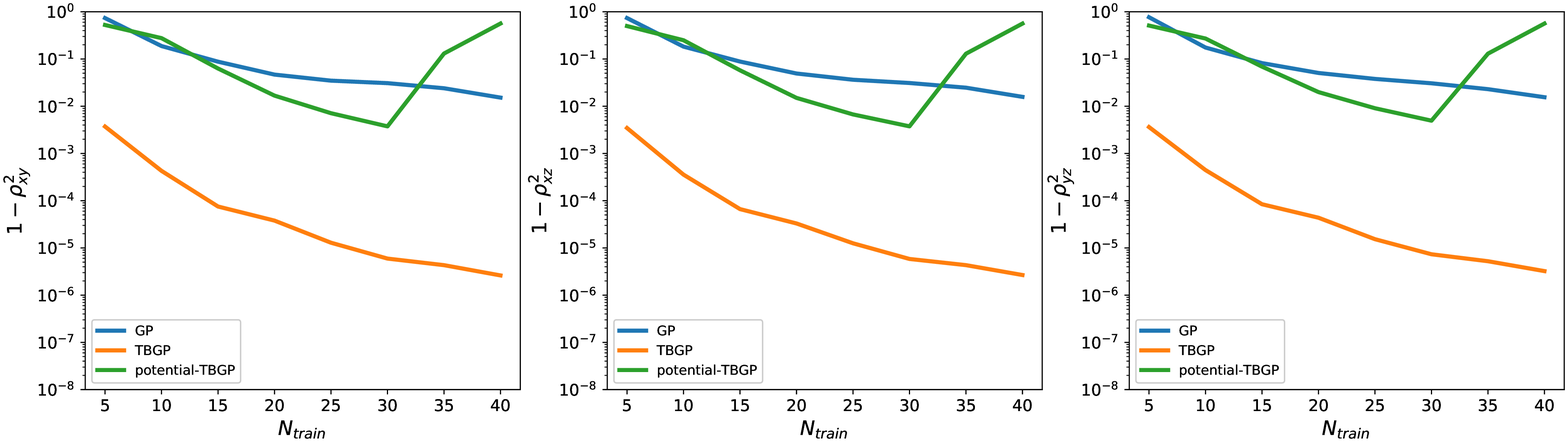}
\caption{The component-wise fraction of variance unexplained ($1-\rho^2$) for the stress tensor predictions as a function of training set size for $[l^2,u^2] = [0.9,2.0]$. Top row are the tension components, and bottom are the shear components.}
\label{fig:stress_corr}
\end{figure}

The results shown in Figures \ref{fig:stress_rmse}, \ref{fig:stress_corr2}, and \ref{fig:stress_corr} are for two datasets, one sampling $[l^2,u^2]=[1.0,1.5]$ and another covering $[0.9,2.0]$. This first choice includes strain values corresponding only to mild extension along the principal axes, while the second choice includes a more extreme range from mild compression to much more extension.
The GP, TBGP, and potential-TBGP formulations were trained on 100 different random samples of datasets of varying size, and each trial's hyperparameters were selected with multi-start L-BFGS optimization \cite{lbfgsb} of the marginal likelihood with 20 random initializations.

The root-mean-square error is shown in Figure \ref{fig:stress_rmse} and fraction of variance unexplained $1-\rho^2$ for correlation coefficient $\rho$ for the stress tensor components are shown in Figures \ref{fig:stress_corr2} and \ref{fig:stress_corr} as a function of the training set size for both datasets.
It is clear that the TBGP has a substantially lower error than the GP with approximately the same rate of convergence, with 1-2 orders of magnitude lower error and 5-6 orders of magnitude lower unexplained variance.
As in the matrix exponential example, the TBGP formulation is uniformly the best choice at all tested numbers of datapoints.
However, the potential-based TBGP has about the same error as the regular GP, and both show slightly degraded performance on predicting the shear components of the stress tensor compared to the tension components.
In addition, for larger sets of data the accuracy of potential-TBGP stops improving and, in the larger deformation case, the error diverges.
We attribute this trend to attempting to learn the full function behavior from gradient information alone, as well as the much larger and more ill-conditioned covariance matrix formed in the inference process.
Individual trials of the training process become more likely to yield higher-error models, increasing the average error.
It is likely that using advanced low-rank factorizations of the covariance matrix or better regularization would reduce the magnitude of this diverging error.

\begin{figure}
\centering
\includegraphics[width=0.45\textwidth]{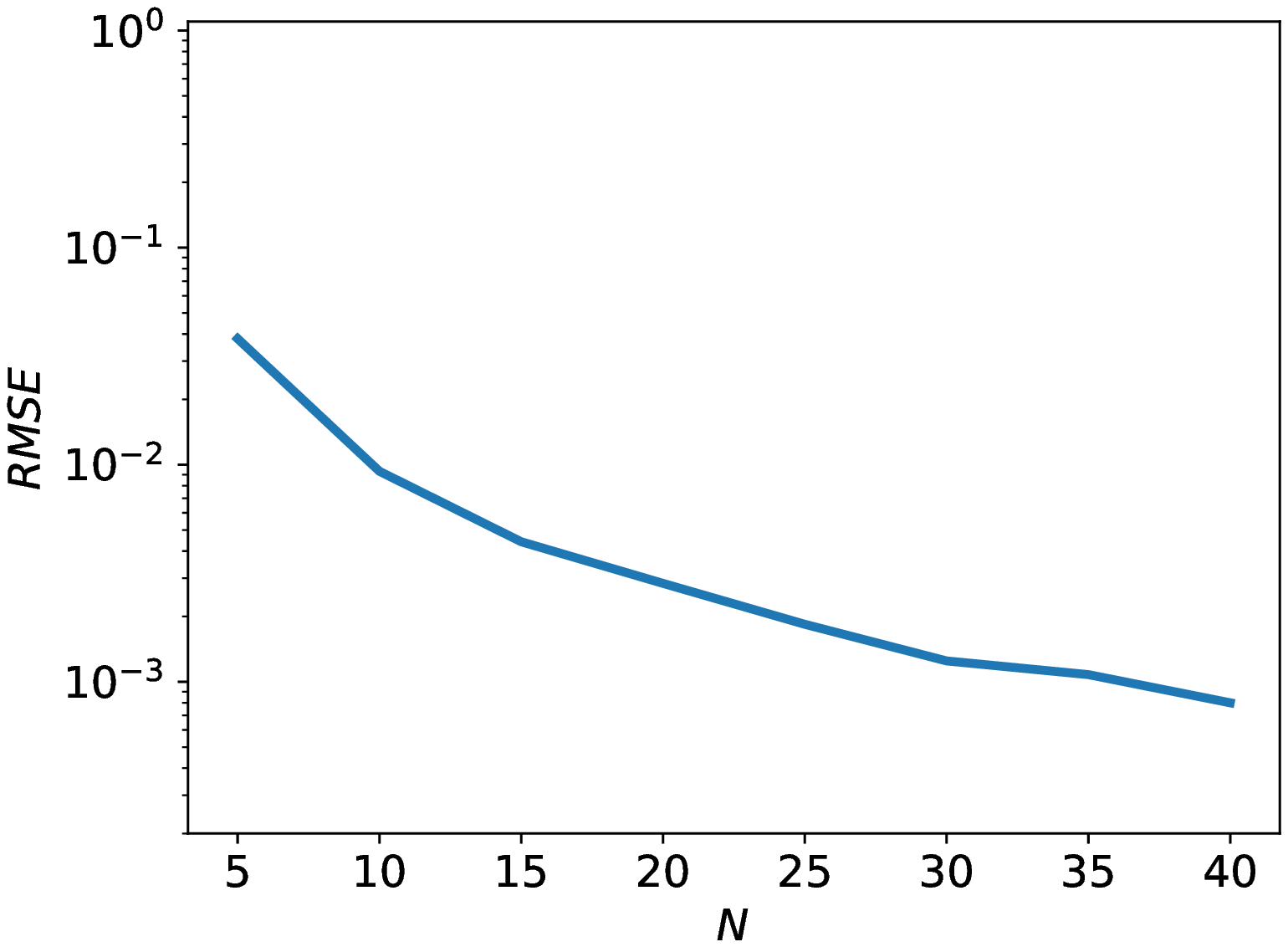}
\includegraphics[width=0.45\textwidth]{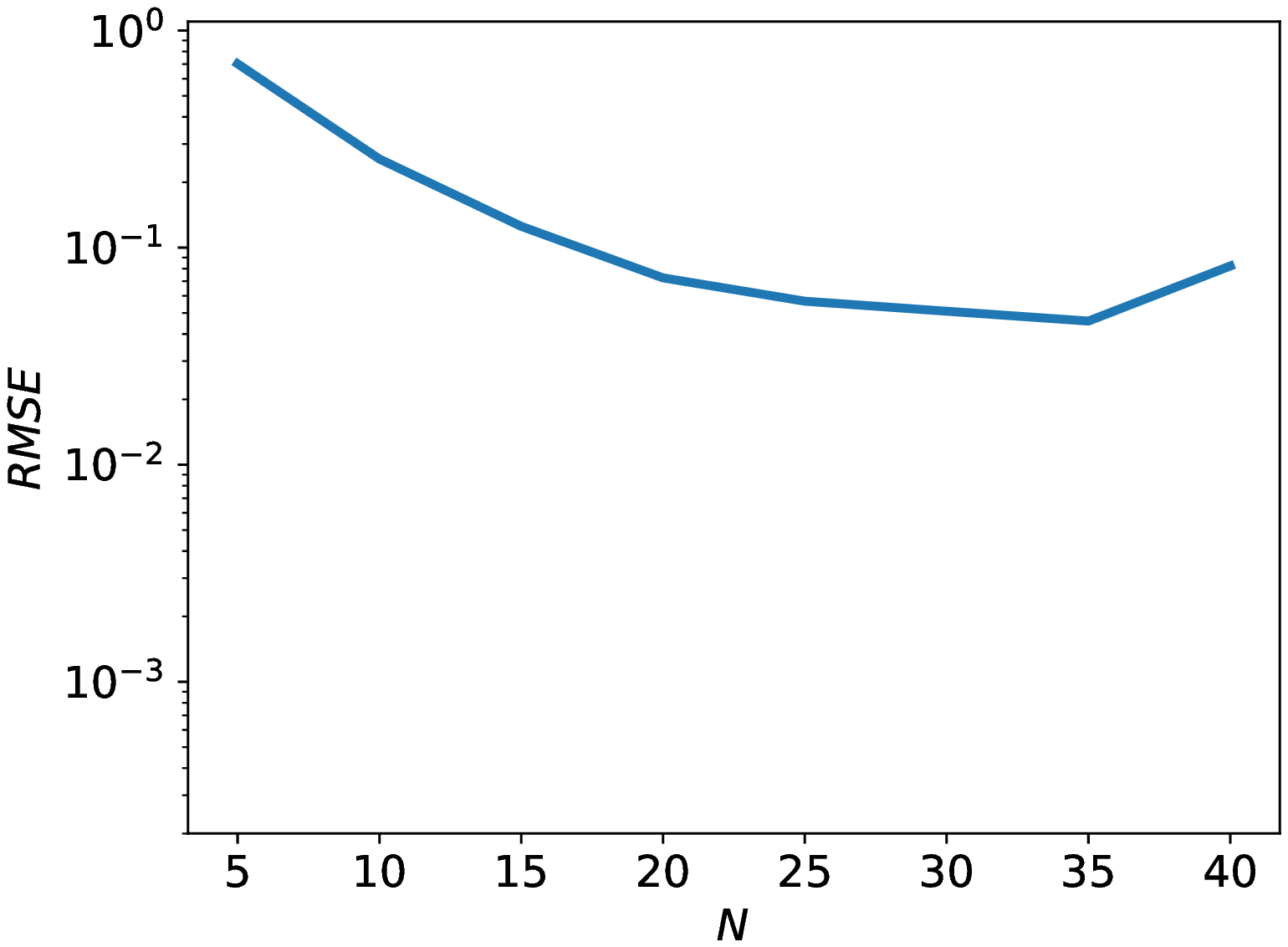}

\includegraphics[width=0.45\textwidth]{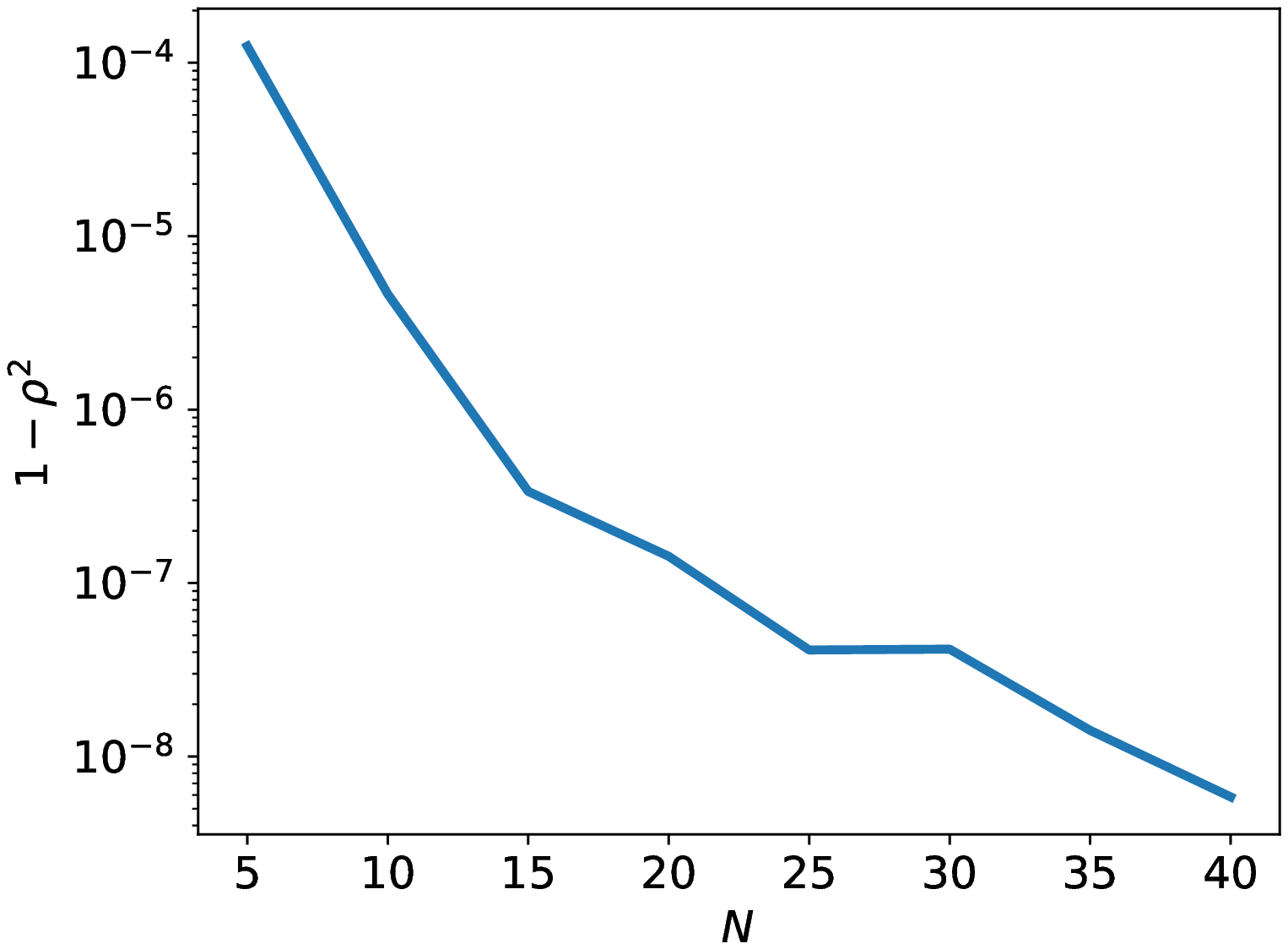}
\includegraphics[width=0.45\textwidth]{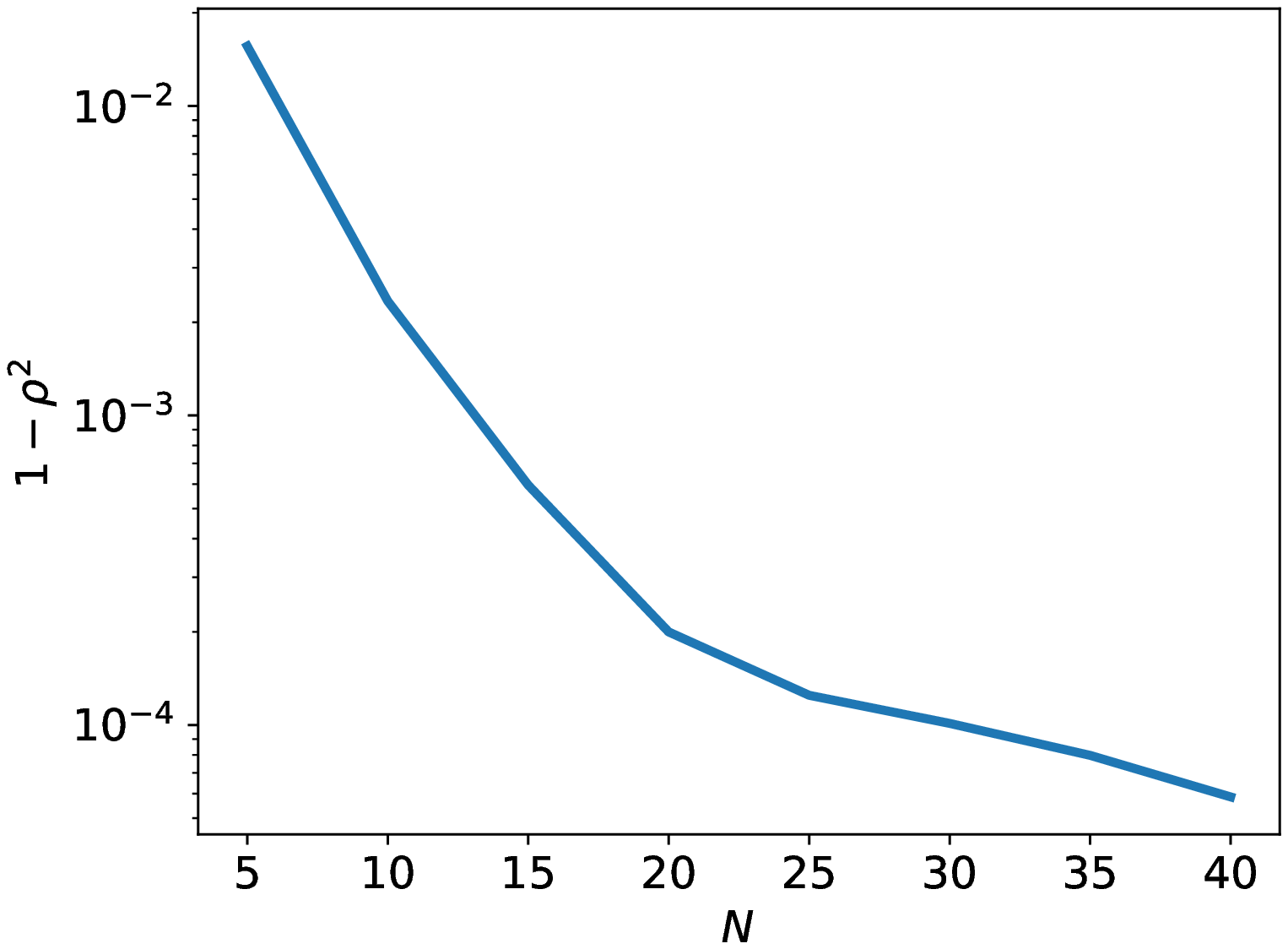}
\caption{The root-mean-squared error (upper) and unexplained variance (lower) of the strain energy density function $\Phi$ as a function of training set size from the potential-TBGP for $[l^2,u^2] = [1.0,1.5]$ (left) and  $[l^2,u^2]=[0.9,2.0]$ (right).
}
\label{fig:pot_err}
\end{figure}

\begin{figure}
\centering
\includegraphics[scale=0.52]{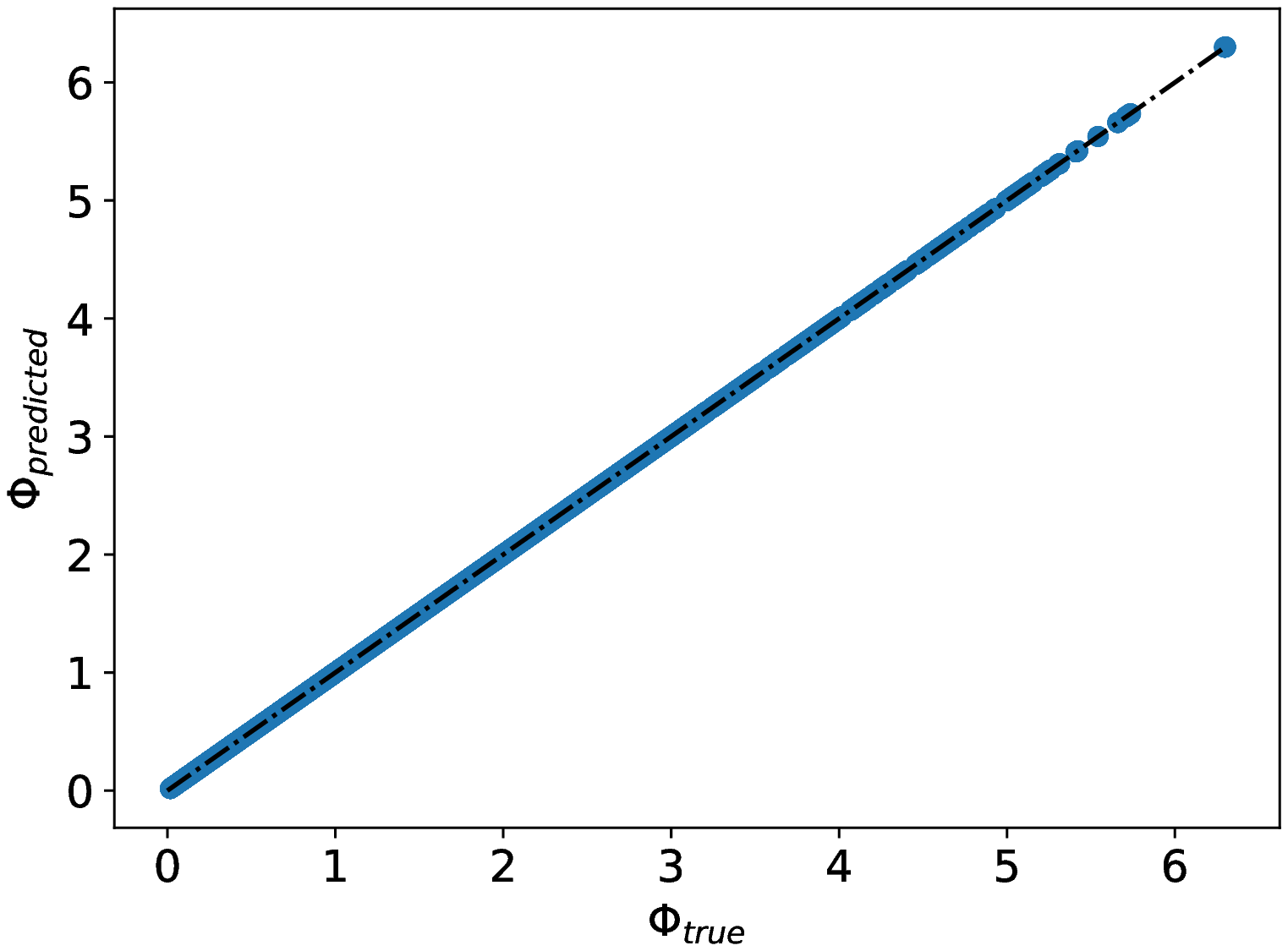}
\includegraphics[scale=0.52]{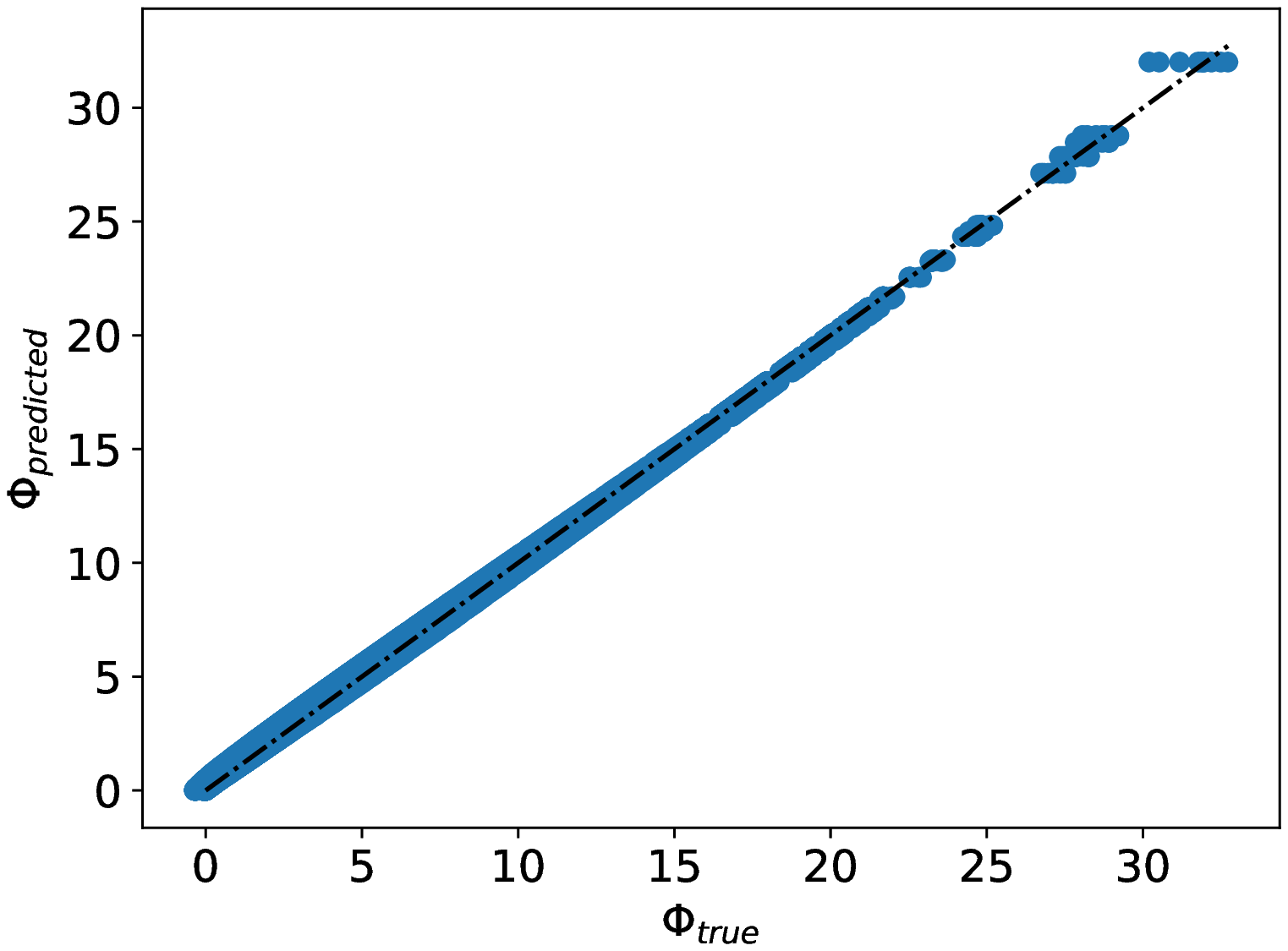}
\caption{Scatter-plot of the predicted and actual strain energy density values at the highest training set size for the cases $[l^2,u^2] = [1.0,1.5]$ (upper panel) and $[0.9,2.0]$ (lower panel).}
\label{fig:pot_compare}
\end{figure}

Although the potential-TBGP performance for predicting the stress components is no better than the GP, it is capable of predicting the strain-energy density function with fairly high accuracy, which is not a task that either of the other two formulations are capable of doing directly.
Figure \ref{fig:pot_err} shows the root-mean-square error and correlation coefficient as a function of training set size. The prediction accuracy does show substantial improvement with increasing training data for the lower-stretch case $[l^2,u^2] = [1.0,1.5]$, but the ill-conditioning issues prevalent in the stress predictions for the higher-stretch case $[l^2,u^2]=[0.9,2.0]$ pervade the potential prediction as well. Figure \ref{fig:pot_compare} shows that the source of disagreement between the potential-TBGP and the Mooney-Rivlin potential originates from values at higher energy, where there is less data and a more complex trend. One possible solution to improve the performance of the potential-TBGP is to use a greedy point selection approach that would use points that span a wider range of invariant space and energy density to train the GP.

%%%%%%%%%%%%%%%%%%%%%%%%%%%%%%%%%%%%%%%%%%%%%%%%%%%%%%%%%%%%%%%%%%%%%%%%%%%
\section{Conclusion}
%%%%%%%%%%%%%%%%%%%%%%%%%%%%%%%%%%%%%%%%%%%%%%%%%%%%%%%%%%%%%%%%%%%%%%%%%%%

This paper has developed and demonstrated an approach to embedding rotational invariance in the Gaussian process regression framework for constitutive modeling of hyperelasticity.
Embedding this physics knowledge led to a dramatic improvement in the accuracy and learning curves for the TBGP formulation compared to the traditional component based approach. 
Also the potential-TBGP formulation demonstrated recovery of the potential function from stress-strain data with comparable accuracy to the plain GP formulation for stress prediction.
While the examples considered here are relatively simple, the application of the methodology to more complex hyper-elastic materials and functions of kinematic variables is straightforward.

One important consideration for future work is the representation of anisotropic material response and functions of multiple tensors.
In these cases, the tensor basis and corresponding invariants are more complex, but the underlying process remains the same.
For example, the second Piola-Kirchoff tensor $\Sb$ may be expressed as a function of the Cauchy-Green deformation tensor $\Cb = \Fb^T \Fb$ and a structure tensor $\Gb$ that characterizes the anisotropy in the response:
\begin{equation}
\Sb = \fb(\Cb,\Gb)
\end{equation}
(see Ref. \cite{zheng1994theory} for more details).
For the case of transverse isotropy where the material response along a direction $\gb$ is different than in the plane perpendicular to the unit vector $\gb$, $\Gb = \gb\otimes\gb$ can be employed as the structure tensor.
The corresponding expansion for $\Sb$ is
\begin{equation}
\Sb = \sum_{i=1}^6 c_i \Ab_i
\end{equation}
in terms of the tensors $\Ab_i \in \{\Ib,\Cb,\Cb^2,\Gb,\Cb\Gb+\Gb\Cb,\Cb^2 \Gb + \Gb\Cb^2\}$ and coefficients $c_i$ which  are functions of the extended invariant set $\{\tr\Cb,\tr\Cb^2, \tr\Cb^3, \tr\Cb\Gb, \tr\Cb^2\Gb\}$ (refer to Ref. \cite{boehler1987representations} and use $\Gb^2 = \Gb$).
Since $\Sb$ is a symmetric tensor and the tensor basis $\{ \Ab_i \}$ is a linearly independent set, the expansion gives 6 equations for the 6 unknowns $c_i$ which may be solved readily.
In this way, the corresponding coefficients as a function of the invariants may be inferred and a TBGP may be trained to make predictions for an anisotropic material response.

There is also room to improve the predictions of the potential-based TBGP. The squared-exponential kernel was selected for its simplicity and smoothness, but it is possible that a more complex or non-stationary kernel would be able to capture the behavior of the potential function in invariant-space more accurately and overcome the ill-conditioning issues seen in this formulation.

%%%%%%%%%%%%%%%%%%%%%%%%%%%%%%%%%%%%%%%%%%%%%%%%%%%%%%%%%%%%%%%%%%%%%%%%%%%
\section*{Acknowledgments}
%%%%%%%%%%%%%%%%%%%%%%%%%%%%%%%%%%%%%%%%%%%%%%%%%%%%%%%%%%%%%%%%%%%%%%%%%%%
This work was supported by the LDRD program at Sandia National Laboratories, and its support is gratefully acknowledged.
Sandia National Laboratories is a multimission laboratory managed and operated by National Technology and Engineering Solutions of Sandia, LLC., a wholly owned subsidiary of Honeywell International, Inc., for the U.S. Department of Energy's National Nuclear Security Administration under contract DE-NA0003525.
The views expressed in the article do not necessarily represent the views of the U.S. Department of Energy or the United States Government. This manuscript has been deemed UUR with SAND number SAND2019-15436J.

%%%%%%%%%%%%%%%%%%%%%%%%%%%%%%%%%%%%%%%%%%%%%%%%%%%%%%%%%%%%%%%%%%%%%%%%%%%
\bibliography{gpr.bib}
%%%%%%%%%%%%%%%%%%%%%%%%%%%%%%%%%%%%%%%%%%%%%%%%%%%%%%%%%%%%%%%%%%%%%%%%%%%

\end{document}